%% file: main.tex
\pdfoutput=1
\documentclass[11pt]{article}

\usepackage{acl}

\usepackage{times}
\usepackage{latexsym}

\usepackage[T1]{fontenc}
\usepackage[utf8]{inputenc}

\usepackage{microtype}

\usepackage{graphicx} %插入图片的宏包
\usepackage{float} %设置图片浮动位置的宏包
\usepackage{subfigure} %插入多图时用子图显示的宏包
\usepackage{bm}
\usepackage{amsmath}
\usepackage{makecell}
\usepackage{booktabs}
\usepackage{amsfonts}

\title{MISC: A \underline{MI}xed \underline{S}trategy-Aware Model\\ Integrating \underline{C}OMET for Emotional Support Conversation}
\author{Quan Tu$^{1}$\thanks{\ \ Equal Contribution.}\ \ \thanks{\ \ This work was done during internship at Xiaomi AI Lab.}\ \ , Yanran Li$^{2}$\footnotemark[1]\ \ , Jianwei Cui$^2$, Bin Wang$^2$, \textbf{Ji-Rong Wen}$^{1,3}$ \and \textbf{Rui Yan}$^{1,3}$\thanks{\ \ Corresponding author: Rui Yan (ruiyan@ruc.edu.cn).} \\
$^1$Gaoling School of Artificial Intelligence, Renmin University of China \\
$^2$Xiaomi AI Lab \\
$^3$Beijing Academy of Artificial Intelligence \\
{\tt $^{1}$\{quantu,jrwen,ruiyan\}@ruc.edu.cn} \\
{\tt $^{2}$\{liyanran,cuijianwei,wangbin11\}@xiaomi.com} \\
}
\begin{document}
\maketitle
\begin{abstract}
%Empathy is critical for emotional support conversation.
Applying existing methods to emotional support conversation---which provides valuable assistance to people who are in need---has two major limitations: (a) they generally employ a conversation-level emotion label, which is too coarse-grained to capture user's instant mental state; (b) most of them focus on expressing empathy in the response(s) rather than gradually reducing user's distress. To address the problems, we propose a novel model \textbf{MISC}, which firstly infers the user's fine-grained emotional status, and then responds skillfully using a mixture of strategy. Experimental results on the benchmark dataset demonstrate the effectiveness of our method and reveal the benefits of fine-grained emotion understanding as well as mixed-up strategy modeling. Our code and data could be found in \url{https://github.com/morecry/MISC}.
\end{abstract} 

\input{1-introduction}
\input{2-related}
\input{3-preliminaries}

\input{4-model}

\input{5-experiments}

\input{6-analysis}

\input{7-conclusion}

\section*{Acknowledgements}
We would like to thank the anonymous reviewers for their constructive comments. This work was supported by National Natural Science Foundation of China (NSFC Grant No. 62122089 \& No. 61876196), Beijing Outstanding Young Scientist Program (NO. BJJWZYJH012019100020098), and Intelligent Social Governance Platform, Major Innovation \& Planning Interdisciplinary Platform for the "Double-First Class" Initiative, Renmin University of China. Rui Yan is the corresponding author, and is supported as a young fellow at Beijing Academy of Artificial Intelligence (BAAI).

% For Arxiv

% \bibliographystyle{acl_natbib}

\appendix
\section{Distribution of Strategies}
\label{appendix:Distribution of Strategy}
\begin{figure}[htbp]
    \centering
    \includegraphics[width=0.5\textwidth]{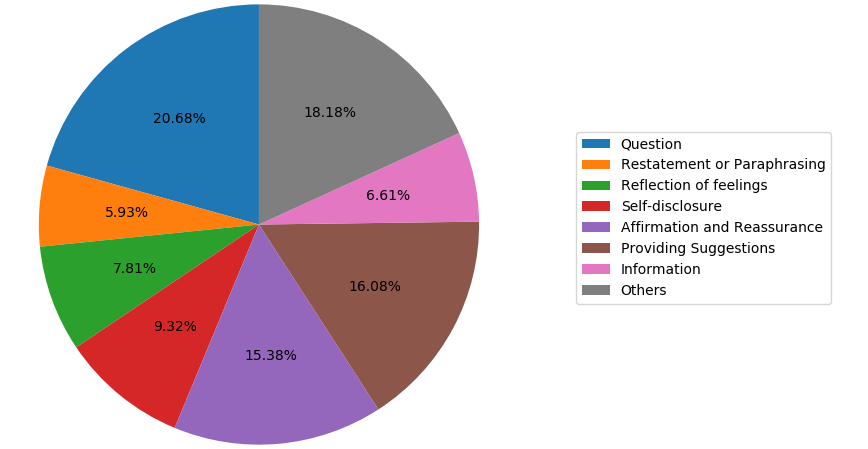}
    \caption{The strategy distribution in the original ESConv dataset.}
    \label{fig:2}
\end{figure}

As show in Figure~\ref{fig:2}, we can see that the proportion of each strategy is relatively balanced. 

\section{Definition of Strategies}
\label{appendix:Definition of Strategies}
Here, we directly adopted from \cite{liu2021towards} to help readers to learn about the specific meaning of each strategy more conveniently.\\
\noindent\textbf{Question} Asking for information related to the
problem to help the help-seeker articulate the issues that they face. Open-ended questions are best, and closed questions can be used to get specific information.\\
\noindent\textbf{Restatement or Paraphrasing} A simple, more concise rephrasing of the help-seeker’s statements that could help them see their situation more clearly.\\
\noindent\textbf{Reflection of Feelings} Articulate and describe the help-seeker’s feelings.\\
\noindent\textbf{Self-disclosure} Divulge similar experiences that you have had or emotions that you share with the help-seeker to express your empathy.\\
\noindent\textbf{Affirmation and Reassurance} Affirm the help-seeker’s strengths, motivation, and capabilities and provide reassurance and encouragement.\\
\noindent\textbf{Providing Suggestions} Provide suggestions about how to change, but be careful to not overstep and tell them what to do.\\
\noindent\textbf{Information} Provide useful information to the help-seeker, for example with data, facts, opinions, resources, or by answering questions.\\
\noindent\textbf{Others} Exchange pleasantries and use other support strategies that do not fall into the above categories.

\section{Description of COMET Relations}
\label{appendix:Definition of Strategies}
In the section, we also adopted the description from~\cite{bosselut2019comet}, so as reader needn't to find it in original text.\\
\noindent\verb|oEffect| The effect the event has on others besides Person X.\\
\noindent\verb|oReact| The reaction of others besides Person
X to the event.\\
\noindent\verb|oWant| What others besides Person X may
want to do after the event. \\
\noindent\verb|xAttr| How Person X might be described
given their part in the event.\\
\noindent\verb|xEffect| The effect that the event would have
on Person X.\\
\noindent\verb|xIntent| The reason why X would cause the
event.\\
\noindent\verb|xNeed| What Person X might need to do before the event.\\
\noindent\verb|xReact| The reaction that Person X would
have to the event.\\
\noindent\verb|xWant| What Person X may want to do after
the event.

\end{document}

%% file: 1-introduction.tex
\section{Introduction}
\label{intro}
Empathy is the ability to perceive what others feel, think in their places and respond properly. It has a broad application scenarios to endow machines with the ability of empathy, including automatic psycho-therapist, intelligent customer service, empathetic conversational agents, and etc~\cite{Fitzpatrick2017DeliveringCB,shin2019happybot,ma2020survey}.

In this work, we focus on a special kind of human-computer empathetic conversation, i.e., emotional support conversation~\cite{liu2021towards}. Distinguishedly, emotional support conversation happens between a seeker and supporter, where the supporter aims to gradually reduce seeker's distress as the conversation goes. This makes existing approaches unsuitable for our setting for at least two reasons. Firstly, existing work on emotional chatting learns to predict user emotion using a conversation-level emotion label, which is \emph{coarse-grained} and \emph{static} to the conversation context~\cite{Rashkin2019TowardsEO,lin2019caire,li2019empdg}. However, emotion is complex and user emotion intensity will \emph{change} during the developing of the conversation~\cite{liu2021towards}. It is thus a necessity to tell seeker's \emph{fine-grained} mental state at each utterance. Secondly, most of empathetic chatbots are trained to respond emotionally in accordance with the predicted coarse-grained emotion class, without consideration on how to address the seeker's emotional problem~\cite{de2012identification,majumder2020mime,xie2021empathetic}. Hence, they are deficient to apply for emotional support conversation whose goal is to help others work through the challenges they face. 

\begin{figure}[!h]
    \centering
    \includegraphics[width=0.45\textwidth]{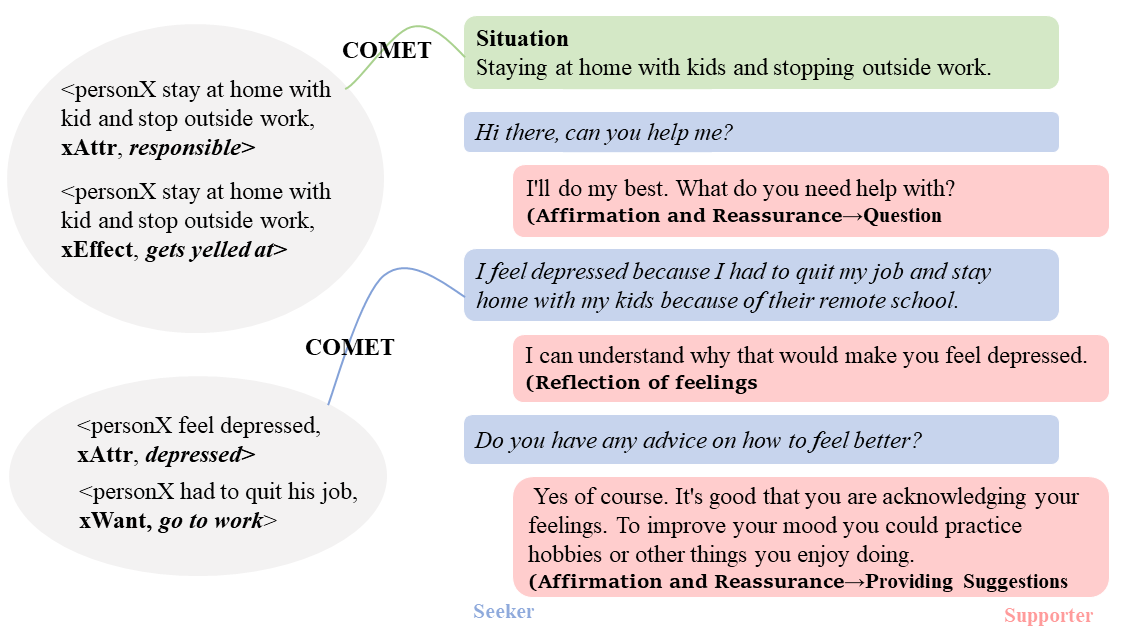}
    \caption{An Emotional Support Conversation Example.}
    \label{fig:1}
\end{figure}

To tackle these issues, we propose a novel approach \textbf{MISC}, a.k.a.  \textbf{MI}xed \textbf{S}rategy-aware model integrating \textbf{C}OMET for emotional support conversation. As to the first issue, we introduce COMET, a pre-trained generative commonsense reasoning model~\cite{bosselut2019comet}, and devise an attention mechanism to selectively adopt the COMET knowledge tuples for fine-grained emotion understanding. As shown in Figure~\ref{fig:1}, this allows us to capture seeker's instantaneous mental state using different COMET tuples. In addition, we propose to also consider response strategy when generating empathetic responses for the second issue. Instead of modeling response strategy as a one-hot indicator, we formulate it as a probability distribution over a strategy codebook, and guide the response generation using a mixture of strategies. At last, our {MISC} produces supportive responses based on both COMET-enhanced mental information and distributed strategy representation. The unique design of mixed strategy not only helps to increase the expressed empathy, but also facilitates to learn the gradual transition in the long response, as the last utterance in Figure~\ref{fig:1}, which will in turn make the conversation more smooth. 

To evaluate our model, we conduct extensive experiments on ESConv benchmark~\cite{liu2021towards} and compare with 5 state-of-the-art empathetic chatbots. Based on both automatic metrics and manual judgments, we demonstrate that the responses generated by our model {MISC} are more relevant and empathetic. Besides, additional experimental analysis reveal the importance of response strategy modeling, and sheds light on how to learn a proper response strategy as well as how response strategy could influence the empathy of the chatbot. % Our code could be found in \url{http://github.com/xxxx}.

In brief, our contributions are as follows: (1) We present a Seq2Seq model {MISC}, which incorporates commonsense knowledge and mixed response strategy into emotional support conversation; (2) We conduct experiments on ESConv dataset, and demonstrate the effectiveness of the proposed {MISC} by comparing with other SOTA methods. (3) We implement different ways of strategy modeling and give some hints on strategy-aware emotional support conversation. 
% We will release code and data after publication. %could be found in \url{http://github.com/release-after-double-blind-review}.

\iffalse
\begin{itemize}
\item We present a Seq2Seq model {MISC}, which incorporates commonsense knowledge and mixed response strategy into emotional support conversation. 
\item We identify the importance of strategy modeling and model a mixture of strategies to enhance the supportive responding.
\item We conduct experiments on ESConv dataset, and demonstrate the effectiveness of the proposed {MISC} by comparing with other SOTA methods. 
\item We implement different ways of strategy modeling and give some hints on strategy-aware emotional support conversation. 
\end{itemize}
\fi

%% file: 2-related.tex
\section{Related Work}
\subsection{Emotion-aware Response Generation}
As suggested in~\citet{liu2021towards}, emotion-aware dialogue systems can be categorized into three classes: emotional chatting, empathetic responding and emotional support conversation. Early work target at emotional chatting and rely on emotional signals~\cite{Li2017DailyDialogAM,ecm,Wei2019EmotionawareCM,zhou-wang-2018-mojitalk,song-etal-2019-generating}. Later, some researchers shift focus towards eliciting user’s specific emotion~\cite{Lubis2018ElicitingPE,Li2020EmoElicitorAO}. %For example, \citet{lin-etal-2019-moel} softly combines the outputs from several empathetic decoders (listeners), each for a certain type of emotion. 
Recent work begin to incorporate extra information for deeper emotion understanding and empathetic responding~\cite{Lin2020CAiREAE,li2019empdg,roller2020recipes}. \citet{mkedg} and \citet{care} exploit ConceptNet to enhance emotion reasoning for response generation. Different from them, our work exploits a generative commonsense model COMET~\cite{bosselut-etal-2019-comet}, which enables us to capture seeker's mental states and facilitates strategy prediction in emotional support conversation.

\subsection{Commonsense Knowledge for NLP} 
Recently, there is a large body of literature injecting commonsense knowledge into various NLP tasks, including classification~\cite{Chen2019IncorporatingSC,Paul2019RankingAS}, question answering~\cite{Mihaylov2018KnowledgeableRE,bauer-etal-2018-commonsense,Lin2019KagNetKG}, story and language generation~\cite{Guan2019StoryEG,ji-etal-2020-language}, and also dialogue systems~\cite{ccm,Zhang2020GroundedCG,mkedg,care}. %One closely related work is commonsense conversational model (CCM)~\cite{Zhou2018CommonsenseKA}, which includes a commonsense knowledge interpreter module and combines knowledge vectors with utterance embeddings before feeding to the encoder. 
These dialogue systems often utilize ConceptNet~\cite{Speer2017ConceptNet5A}, aiming to complement conversation utterances with physical knowledge. Distinguished from ConceptNet, ATOMIC~\cite{atomic} covers social knowledge including event-centered causes and effects as well as person-related mental states. To this end, ATOMIC is expected beneficial for emotion understanding and contributing to response empathy. In this work, we leverage COMET~\cite{bosselut-etal-2019-comet}, a commonsense reasoning model trained over ATOMIC for emotional support conversation.
\subsection{Strategy-aware Conversation Modeling}
Conversation strategy can be defined using different notions from different perspectives. A majority of research works is conducted under the notion of dialog acts, where a plethora of dialog act schemes have been created~\cite{Mezza2018ISOStandardDD,Paul2019TowardsUD,Yu2021MIDASAD}. Dialog acts are empirically validated beneficial in both task-oriented dialogue systems and open-domain social chatbots~\cite{Zhao2017LearningDD,Xu2018TowardsEA,Peng2020FewshotNL,hierarchicalTKDE}. As to empathetic dialogues, conversation strategy is often defined using the notion of response intention or communication strategy, which is inspired from the theories of empathy in psychology and neuroscience~\cite{Lubis2019DialogueMA,Li2021TowardsAO}. Whereas \citet{Welivita2020ATO} define a taxonomy of 15 response intentions through which humans empathize with others, \citet{liu2021towards} define a set of 8 support strategies that humans utilize to reduce other's emotional distress. This partially reveals that response strategy is complex, which motivates us to condition on a mixture of strategy when generating supportive responses.

%% file: 3-preliminaries.tex
\section{Preliminaries}
\subsection{ESConv Dataset}
In this paper, we use the \textbf{E}motional \textbf{S}upport \textbf{Conv}ersation dataset, \textbf{ESConv}~\cite{liu2021towards}. Before conversations start, seekers should determine their emotion types, and tell the situation they are dealing with to supporters.  
% During the conversation, the seeker should give feedback~(point 1~5) based on their satisfaction toward the supporter's utterance. 
Besides, the strategy of every supporter's utterance is marked, which is the most important to our work. In total, there are 8 kinds of strategies, and they are almost evenly distributed. More details are given in Appendix. % strategy distributions in the original dataset is given in the appendix. % Figure~\ref{fig:2}.
\iffalse
\begin{figure}[!h]
    \centering
    \includegraphics[width=0.5\textwidth]{figures/strategy-dist.png}
    \caption{The strategy distribution in the original ESConv dataset.}
    \label{fig:2}
\end{figure}
\fi

\subsection{Problem Formulation}
For general dialogue response generation, the target is to estimate the probability distribution $p(\bm{r}|\bm{c})$ of the dataset $\mathcal{D}=\{\bm{c}^{(i)}, \bm{r}^{(i)}\}_{i=1}^{N}$, where $\bm{c}^{(i)}=(\bm{u}_1^{(i)}, \bm{u}_2^{(i)},...,\bm{u}_{n_i}^{(i)})$ consists of a sequence of $n_i$ utterances in the dialogue history, and $\bm{r}^{(i)}$ is the target response. For the sake of brevity, we omit the superscript $(i)$ when denoting a single example in the remaining part. %of the manuscript. Unless otherwise stated, all notations without the superscript $(i)$ can be regarded as a single sample.

% &=[\bm{b}_{st, 1}^{(i)}, \bm{b}_{st, 2}^{(i)}, \bm{b}_{st, N_{st}}^{(i)}] \\

In the setting of emotional support conversation, the seeker's situation $\bm{s}$ is considered as an extra input, which describes the seeker's problem in free-form text. We also denote the seeker's last post (utterance) as $\bm{x}$. Consequently, the target becomes to estimate the probability distribution $p(\bm{r}|\bm{c, s, x})$. %Here, $\bm{x}^{(i)}=\bm{u}_{n_p}^{(i)}$, ${n_p}$ is the index of the seeker's last post in $\bm{C}^{(i)}$.

%% file: 4-model.tex
\section{Model: MISC}
\begin{figure*}[htpb]
    \centering
    \includegraphics[width=1.0\textwidth]{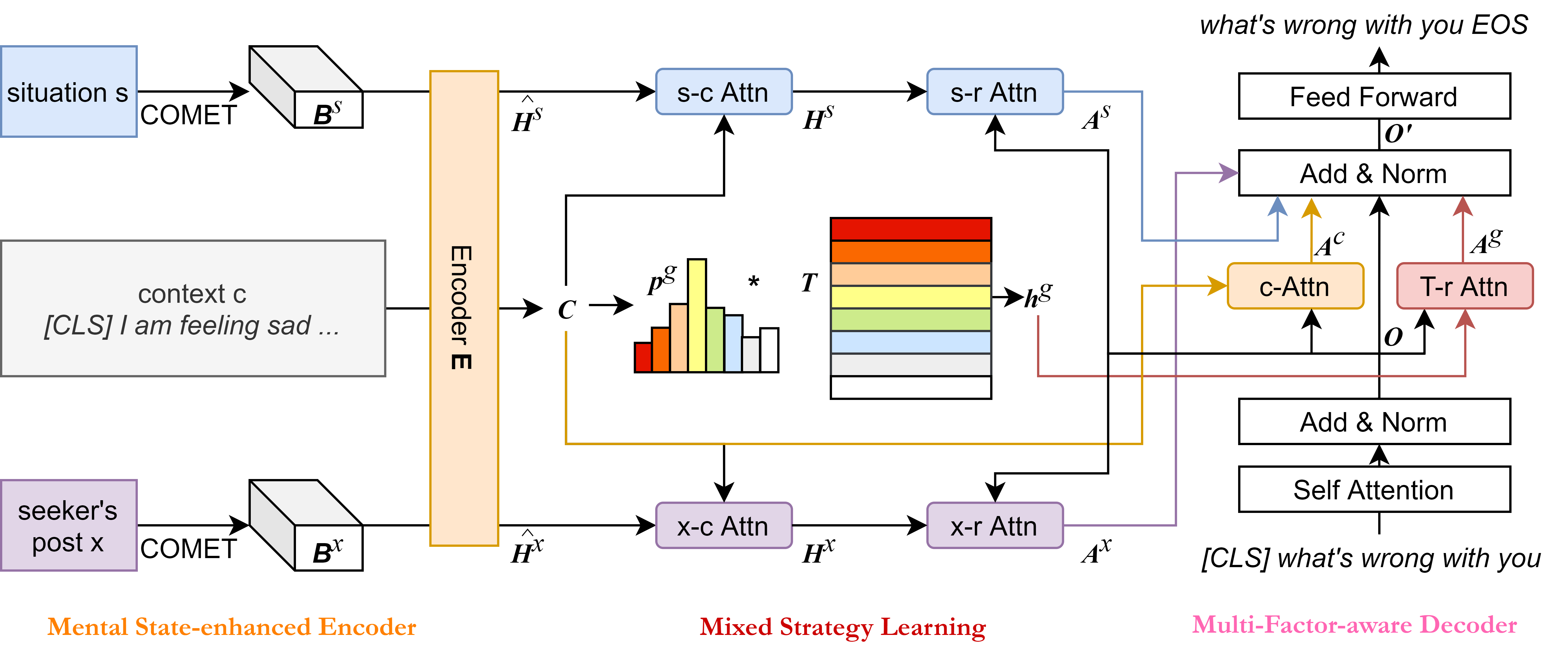}
    \caption{The overview of the proposed MISC which consists of a mental state-enhanced encoder, a mixed strategy learning module, and a multi-factor-aware decoder.}
    \label{fig:3}
\end{figure*}

The overview of our approach is shown in Figure~\ref{fig:3}. Based on blenderbot-small~\cite{roller2020recipes}, our model {MISC} consists three main components: (1) a mental state-enhanced encoder~\cite{bosselut2019comet}; (2) a mixed strategy learning module; and (3) a multi-factor-aware decoder. %In the following sections, we will describe each component in detail.

\subsection{Mental State-Enhanced Encoder}
Following common practice, we firstly represent the context using the encoder $\mathtt{E}$: 
\begin{equation}
    \bm{C} = \bm{\mathtt{E}}(\mathtt{CLS}, \bm{u}_1,\mathtt{EOS}, \bm{u}_2,...,\bm{u}_{n_i})
    \label{eq: context encode}
\end{equation}
where $\mathtt{CLS}$ is the start-token and $\mathtt{EOS}$ is the separation-token between two utterances.

To better understand the seeker's situation, we exploit COMET~\cite{bosselut2019comet}, a commonsense knowledge generator to supply mental state information related to the conversation. Concretely, we treat the situation $\bm{s}$ as an event, and feed it with different relations into COMET:
\begin{equation}
    \bm{B}^{s} = \bigcup\limits_{j=1}^{N_r}\mathtt{COMET}(rel_j, \bm{s})
\end{equation}
% \begin{equation}
%     \bm{BP}^{(i)} = \bigcup\limits_{j=1}^{N_r}\mathrm{COMET}(rel_j, \bm{lp}^{(i)})
% \end{equation}
where $N_r$ is the number of pre-defined relations in COMET, and $rel_j$ stands for the $j$-th specific relation, such as \verb|xAttr| and \verb|xReact|.\footnote{Please refer to the appendix file for the definitions of all the relations as well as a brief introduction of COMET.} Note that given a certain event-relation pair, COMET is able to generate multiple ``tails'' of free-form mental state information, $\bm{B}^{s}$ is a set of $N_{s}$ mental state blocks, i.e., $\bm{B}^{s}=\{\bm{b}^s_{j}\}_{j=1}^{N_s}$. Similarly, we can obtain the set of mental state blocks $\bm{B}^{x}$ using the seeker's last post $\bm{x}$.

Then, all of the free-form blocks will be transformed into dense vectors using our encoder $\mathtt{E}$:
\begin{equation}
    \begin{aligned}
    \hat{\bm{H}}^{s}&=[\bm{h}^s_{1,1}, \bm{h}^s_{2,1},...,\bm{h}^s_{N_{st},1}]\\
    \bm{h}^s_{j}&=\mathtt{E}(\bm{b}^s_{j})
    \end{aligned}
\end{equation}
and the hidden state of each block's first token will be used to represent the corresponding block. Later, due to the noisy of COMET blocks, a lot of them are irrelevant to the context. We creatively take attention method to refine the strongly relevant blocks. That operation could be expressed as 
\begin{equation}
    \begin{aligned}
    \bm{Z}&= \mathtt{softmax}(\hat{\bm{H}}^{s}\cdot\bm{C}^{\mathrm{T}})\cdot\bm{C}\\
   \bm{H}^{s}&=\mathtt{LN}(\hat{\bm{H}}^{s}+\bm{Z})
    \end{aligned}
\end{equation}
where $\mathtt{LN}$ is the LayerNorm module~\cite{ba2016layer}. Similarly, we could transform $\bm{x}$ to $\bm{H}^{x}$ following the same method as $\bm{s}$ to $\bm{H}^{s}$. At last, we get the conversation-level and utterance-level representation of seeker's mental state $\bm{H}^{s}$ and $\bm{H}^{x}$, which are enhanced with commonsense information.

\subsection{Mixed Strategy Learning Module}
One straightforward way to predict the response strategy is to train a classifier upon the $\mathrm{CLS}$ states of the context representation $\bm{C}$ from Eq.~(\ref{eq: context encode}):
\begin{equation}
   \bm{p}^{g}=\mathtt{MLP}(\bm{C}_1)
\end{equation}
where $\mathtt{MLP}$ is a multi-layer perceptron, and $\bm{p}^{g}$ records the probabilities of each strategy to be used. 

To model the complexity of response strategy as discussed before, we propose to employ the distribution $\bm{p}^{g}$ and model a mixture of strategies for response generation. 
% Inspired by VQ-VAE~\cite{oord2017neural}, we introduce a strategy codebook $\bm{T} \in  \mathcal{R}^{m\times d}$ as an embedding table, which represents $m$ strategy latent vectors with the dimension $d$.
Here, we masterly learn from the idea of VQ-VAE’s codebook to represent strategy\cite{oord2017neural}. The strategy codebook $\bm{T} \in \mathbb{R}^{m\times d}$ represent $m$ strategy latent vectors (here $m$ = 8) with the dimension size $d$.  By weighting $\bm{T}$ using $\bm{p}^{g}$, we are able to obtain a comprehensive strategy representation $\bm{h}^{g}$ 
\begin{equation}
   \bm{h}^{g}=\bm{p}^{g}\cdot\bm{T}
\end{equation}

Our codebook-based method has two benefits: (1) It is beneficial when long responses are needed to skillfully reduce the seeker's distress, which is common in emotional support conversation. (2) It is flexible to learn. Intuitively, if a strategy has a higher probability in $\bm{p}^{g}$, it should take greater effect in guiding the support conversation. %It is also compatible with single strategy learning. 
In the extreme case where we have a sharp distribution, one single strategy will take over the control. 
% (3) By using the codebook, each strategy is assigned with a latent vector, which could be learned separately and induced with extra learning objectives. Hence, our way can be regarded as disentangling strategy learning from response generation. The benefits of our codebook-based mixed strategy learning will be carefully analyzed in Section~\ref{sec:analysis}. 
%The effect will be analyzed in It can be regarded as a kind of disentanglemen   $\bm{h}^{g}$ is assumed to 
%But we hold the point that single strategy vector is not proper for the response because a response is usually consist of multiple sentences which take different strategies. 

\subsection{Multi-Factor-Aware Decoder}
The remaining is to properly utilize the inferred mental states and the strategy representation. To notify the decoder of these information, we modify the backbone's cross attention module as:
\begin{equation}
    \begin{aligned}
        \bm{A}^{c}&=\mathtt{CROSS\text{-}ATT}(\bm{O}, \bm{H}) \\
        \bm{A}^{s}&=\mathtt{CROSS\text{-}ATT}(\bm{O}, \bm{H}^{s})\\
        \bm{A}^{x}&=\mathtt{CROSS\text{-}ATT}(\bm{O}, \bm{H}^{x}) \\ 
        \bm{A}^{g}&=\mathtt{CROSS\text{-}ATT}(\bm{O}, \bm{h}^{g}) \\
        \bm{O}^{'}&=\mathtt{LN}(\bm{A}^{c} + \bm{A}^{s} + \bm{A}^{x} + \bm{A}^{g} + \bm{O})
    \end{aligned}    
\end{equation}
where $\mathtt{CROSS\text{-}ATT}$ stands for the backbone's cross attention module, and $\bm{O}$ is the hidden states of the decoder, which produces the final response by interacting with multi-factors. %the fine-grained mental state and the mixed response strategy.

Based on blenderbor-small~\cite{roller2020recipes}, we jointly train the model to predict the strategy and produce the response: 
\begin{equation}
    \begin{aligned}
        \mathcal{L}_r &= -\sum\limits_{t=1}^{n_r}\mathtt{log}(p(r_t|\bm{r}_{j<t},\bm{c},\bm{s}, \bm{x})) \\
        \mathcal{L}_g &=-\mathtt{log}(p(g|\bm{c},\bm{s}, \bm{x}))\\
        \mathcal{L} &= \mathcal{L}_r + \mathcal{L}_{g}
    \end{aligned}
\end{equation}
where $n_r$ is the length of response, $g$ is the true strategy label, $\mathcal{L}_{g}$ is the loss of predicting strategy, $\mathcal{L}_r$ is the loss of predicting response, and $\mathcal{L}$ is combined objective to minimize.

%% file: 5-experiments.tex
\section{Experiments}
\subsection{Experimental Setups}
We evaluate our and the compared approaches on the dataset ESConv~\cite{liu2021towards}. For pre-processing, we truncate the conversation examples every 10 utterances, and randomly spilt the dataset into train, valid, test with the ratio of 8:1:1. The statistics is given in Table~\ref{tab:1}.

\begin{table}[htbp]
\centering
\small
\begin{tabular}{lccc} 
\toprule
\textbf{Category}& \textbf{Train} & \textbf{Dev} & \textbf{Test}\\
\midrule
\# dialogues & 14117& 1764&1764 \\
Avg. \# words per utterance & 17.25& 17.09&17.11 \\
Avg. \# turns per dialogue & 7.61& 7.58&7.49 \\
Avg. \# words per dialogue & 148.46& 146.66&145.17 \\
\bottomrule
\end{tabular}
\caption{The statistics of processed ESConv dataset.}
\label{tab:1}
\end{table}

\subsection{Evaluation Metrics}
We adopt a set of automatic and human evaluation metrics to assess the model performances: 
% It's notable that although \citet{sharma2020computational}  proposed 3 automatic empathy scores, they are trained using a small set of Reddit posts, which are very different from the setting of emotional support conversation. We have tried these scores but found they are very inconsistent with human evaluation results. As such, they are not relible indicators in this work.

\noindent\textbf{Automatic Metrics}. (1) We take the strategy prediction accuracy \textbf{ACC.} as an essential metric. A higher ACC. indicates that the model has a better capability to choose the response strategy. (2) We then acquire the conventional \textbf{PPL} (perplexity), \textbf{B-2} (BLEU-2), \textbf{B-4} (BLEU-4)~\cite{papineni2002bleu}, \textbf{R-L} (ROUGE-L)~\cite{lin2004rouge} and \textbf{M} (Meteor)~\cite{denkowski2014meteor} metrics to evaluate the lexical and semantic aspects of the generated responses. (3) For response diversity, we report \textbf{D-1} (Distinct-1) and \textbf{D-2} (Distinct-2) numbers, which assesses the ratios of the unique n-grams in the generated responses~\cite{li2015diversity}.

%These annotations are then then used to train three empathy classifiers, and the classifier outputs suggest to what degree a given response is empathetic. Based on the classifier results, we are able to automatically quantify the strategy learning result of each emo based on the three numbers. 

\noindent\textbf{Human Judgments}. Following~\citet{see-etal-2019-makes}, we also recruit 3 professional annotators with linguistic and psychologist background and ask them to rate the generated responses %in both point-wise and pair-wise settings. (1) For point-wise setting, we adopt Sensibleness and Specificity Average (SSA) score~\cite{Adiwardana2020TowardsAH} to assess whether the given generated response makes sense, and if yes, whether it is specific and informative to the conversation context. %The SSA score has empirically shown better aligned with human annotations, and we use the average SSA score from the three annotators. 
according to Fluency, Knowledge and Empathy aspects with level of \{0,1,2\}. %, which measures to what degree the generated response(s) are fluent, informative and empathetic, respectively. %(2) For pair-wise setting, we pass the query and two responses to the annotators, and ask them to compare which is better. The majority choice will be kept as the final result. 
For fair comparison, the expert annotators do not know which model the response is from. Note that these 3 writers are paid and the results are proof-checked by 1 additional person.% to ensure the quality. %Eventually, the Fleiss' Kappa score for the human agreement is xx (\textbf{to be replacement}), indicating a moderate agreement.

\begin{table*}[htbp]
\centering
\small
%\vspace{-1mm}
%\resizebox{0.45\textwidth}{!}{

\begin{tabular}{lcccccccc}
\toprule
\textbf{Model} &\textbf{ACC}(\%)\,$\uparrow$ & \textbf{PPL}\,$\downarrow$ & \textbf{D-1}\,$\uparrow$ & \textbf{D-2}\,$\uparrow$ &\textbf{B-2}\,$\uparrow$ &\textbf{B-4}\,$\uparrow$& \textbf{R-L}\,$\uparrow$& \textbf{M}(\%)\,$\uparrow$\\ 
\midrule
Transformer &-&89.61&1.29&6.91&6.53&1.37&15.17&10.33 \\
MT Transformer &-&89.52&1.28&7.12&6.58&1.47&14.75&10.27 \\
MoEL &-&133.13&2.33&15.26&5.93&1.22&14.65&9.75\\
MIME &- &47.51&2.11&10.94&5.23&1.17&14.74&9.49 \\
BlenderBot-Joint&28.57&18.49&4.12&17.72&5.78&1.74&16.39&9.93\\
MISC&\textbf{31.63}&\textbf{16.16}&\textbf{4.41}&\textbf{19.71}&\textbf{7.31}&\textbf{2.20}&\textbf{17.91}&\textbf{11.05} \\
\bottomrule
\end{tabular}%}
%\vspace{+0mm}
%\vspace{-4mm}
\caption{Automatic Evaluation Results on ESConv.} %$\uparrow$ means the higher score is better, and $\downarrow$ means the opposite.}
\label{tab:main-exp}
\end{table*}

\begin{table}
\centering
\small

\begin{tabular}{lccc}
\toprule
\textbf{Model} &  \textbf{Flu.} &\textbf{Know.} & \textbf{Emp.}  \\
\midrule
Transformer & 0.62& 0.31& 0.29\\
MT Transformer & 0.78 & 0.34 & 0.82\\
MoEL & 0.36 & 0.80 & 0.33\\
MIME & 1.13 & 0.27 & 0.35\\
BlenderBot-Joint & \textbf{1.87} & 0.74 & 1.21\\
MISC & 1.84 & \textbf{1.06} & \textbf{1.44}  \\
\bottomrule

\end{tabular}
\caption{Manual Evaluation Results. The Fleiss Kappa score~\cite{Fleiss1973TheEO} reaches 0.445, indicating a moderate level of agreements.}
\label{tab:exp_generation_human}
\end{table}

\subsection{Compared Models}
\noindent\textbf{Transformer} is a vanilla Seq2Seq model trained based on the MLE loss~\cite{vaswani2017attention}. 

\noindent\textbf{MT Transformer} is the \textbf{M}ulti-\textbf{T}ask transformer which considers emotion prediction as an extra learning task~\cite{rashkin2018know}. In specific, we use the conversation-level emotion label provided in ESConv to learn emotion prediction.  

\noindent\textbf{MoEL} softly combines the output states from multiple listeners (decoders) to enhance the response empathy for different emotions~\cite{lin2019moel}.

\noindent\textbf{MIME} considers the polarity-based emotion clusters and emotional mimicry for empathetic response generation~\cite{majumder2020mime}. %Similarly, we also use the conversation-level emotion label in the original dataset to train the model.

\noindent\textbf{BlenderBot-Joint} is the SOTA model on ESConv dataset, which prepends a special strategy token before the response utterances~\cite{liu2021towards}.

\subsection{Implementation Details}
We implement our approach based on blenderbot-small~\cite{roller2020recipes} using the default sizes of vocabulary and the hidden states. For the last post $\bm{x}$ and the situation $\bm{s}$, we
% use 9 relations \texttt{xIntent}, \texttt{xEffect}, \texttt{xAttr}, \texttt{xNeed}, \texttt{xReact}, \texttt{xWant}, \texttt{oWant}, \texttt{oReact} and \texttt{oEffect} to inference it by COMET, and
set the maximum number of the retrieved COMET blocks as 30 and 20 respectively. The inferred COMET blocks will be sent to the encoder with a maximum of 10 words.

To be comparable with the SOTA model in~\citet{liu2021towards}, we fine-tune {MISC} based on the blenderbot-small with the size of 90M parameters by a Tesla-V100 GPU. The batch size of training and evaluating is 20 and 50, respectively. We initialize the learning rate as 2e-5 and change it during training using a linear warmup with 120 warmup steps. We use AdamW as optimizer~\cite{loshchilov2018fixing} with $\beta_1$=0.9, $\beta_2$=0.999 and $\epsilon$=1e-8. After training 8 epochs, the checkpoint with the lowest perplexity on the validation set is selected for testing. Following~\cite{liu2021towards}, we also adopt the decoding algorithms of Top-$p$ and Top-$k$ sampling with $p$=0.3, $k$=30, temperature $\tau$=0.7 and the repetition penalty 1.03. We will release the source code to facilitate future work.

\subsection{Experimental Results}
As shown in Table~\ref{tab:main-exp}, the vanilla Transformer performs the worst according to its relatively low PPL, BLEU-n and distinct-n scores. This is not suprising because it does not have any other specific optimization objective to learn the ability of empathy, and it is observed to be deficient for capturing long context as that in the ESConv dataset. 

The performances of MT Transformer, MoEL and MIME, are also disappointing. Even though they three are equipped with empathetic objectives such as emotion prediction and ensembling listener, they are based on the conversation-level static emotion label, which is not adequate for fine-grained emotion understanding. More importantly, these three empathetic models lack of the ability of strategically consoling the seekers in the setting of emotional support conversation. 

By comparing with the SOTA model {BlenderBot-Joint}, we can see that our model {MISC} is more effective especially in predicting more accurate response strategy. Whereas {BlenderBot-Joint} predicts one single strategy at the first decoding step, our method {MISC} models mixed response strategies using a strategy codebook and allows the decoder to learn the smooth transition and exhibit empathy more naturally. The comparison result suggests that it is beneficial to predict the response strategy as an extra task and to take into consideration the strategy complex for emotional support conversation. 

The human evaluation results in Table~\ref{tab:exp_generation_human} are consistent with the automatic results. Thanks to the pretrained LM blenderbot-small~\cite{rashkin2018know}, {BlenderBot-Joint} and our MISC significantly outperform other models on the Fluency aspect. Notably, our MISC yields the highest Knowledge score, which indicates that the responses produced by our approach contain much more specific information related to the context. We conjecture that our multi-factor-aware decoder successfully learns utilize the mental state knowledge from COMET with the mixture of the predicted strategies. 

Overall speaking, {MISC} performs the best on almost every metric. It strongly demonstrates the effectiveness of our approach, and highlights the importance of fine-grained mental state modeling and mixed response strategy incorporation.

%% file: 6-analysis.tex
\section{Analysis}
Our method {MISC} has two novel designs: considering the fine-grained mental states and incorporating a mixture of response strategy. To investigate more, we conduct extra experiments, % to answer the following questions: (1) How much does each part of our {MISC} contributes to the overall performance? (2) What is the difference of the generation results between of our model and other compared model? (3) Is it necessary to perceive seekers' fine-grained mental states instead of the coarse-grained emotion type? (4) How does the mixed strategy works? 
and the analysis results give us hints of how to develop better emotional support conversational agents. % By a few experimental evidence, we will attempt to  these questions one by one.

\subsection{Ablation Study}
In order to verify the improvement brought by each added part ($g$, $s$, $x$), we drop these three parts from the MISC and check the performance changes. As shown in Table~\ref{tab:ablation-exp}, the scores on all the metrics decrease dramatically when the $g$ is albated. Consequently, we suppose the strategy attention is vital for guiding the semantics of the response. In addition, the scores also decline when we remove the the situation $s$ and the seeker's last query $x$. According to the above experiments, each main part of the MISC is proven effective.
\begin{table}[!h]
\centering
\small
%\vspace{-1mm}
\begin{tabular}{lcccc}
\toprule
\textbf{Model}& \textbf{D-1}\,$\uparrow$&\textbf{B-2}\,$\uparrow$ & \textbf{R-L}\,$\uparrow$ &\textbf{M}(\%)\,$\uparrow$\\ 
\midrule
MISC&\textbf{4.41}&\textbf{7.31}&\textbf{17.91}& \textbf{11.05}\\
w/o $g$ &3.85&7.09&16.75&9.85 \\
w/o $s$&4.39&6.35&17.05&10.06\\
w/o $x$&4.27&6.49&17.03&10.09\\
\bottomrule
\end{tabular}
\caption{Evaluation Results of Ablation Study.}
\label{tab:ablation-exp}
\end{table}

% \subsection{Ablation Study}
% \begin{table}[htbp]
% \centering
% %\small
% %\vspace{-1mm}
% \caption{Result of ablation study.}
% \label{tab:ablation-exp}
% \resizebox{0.45\textwidth}{!}{
% \begin{tabular}{lccccccc}
% \toprule
% Model&B-2\,$\uparrow$ & R-L\,$\uparrow$& D-1\,$\uparrow$ & ER\,$\uparrow$ & IP\,$\uparrow$ & EX\,$\uparrow$&\\ 
% \midrule
% MISC&\textbf{7.31}&\textbf{17.91}&\textbf{4.41}&0.71&0.48&0.43& \\
% w/o $sg$ &7.09&16.75&3.85&\textbf{0.83}&\textbf{0.73}&0.17& \\
% w/o $st$ &	6.49&17.03&	4.27&0.66&0.41&\textbf{0.55}\\
% w/o $lp$ &6.35&17.05&4.39&0.69&0.46&0.46\\
% \bottomrule
% \end{tabular}}

%\vspace{+0mm}
%\vspace{-4mm}

\subsection{Case Study}
In Table~\ref{tab:generate example}, an example is present to compare the response generated by the MISC and the other models. Various problems appear in the compared models, such as inconsistency, repetition, contradiction, etc. Intuitively, our model achieves the best performance in contrast. Besides, we present a visualization in Figure~\ref{fig:visualization of the relation} to interpret how the MISC organizes the response under the combined effect of the COMET blocks and the mixture of strategies. 

\begin{table*}[htbp]
\centering
\small

\begin{tabular}{|l|c|}
\hline
\multicolumn{2}{|c|}{\textbf{Situation}}\\
\hline
Seeker &\makecell[l]{My boyfriend and I recently broke up due to long-distance relationship and the impact COVID-19...}\\%%has had on the travel industry.}\\
\hline
\multicolumn{2}{|c|}{\textbf{Context}}\\
\hline
Seeker &\makecell[l]{Hi, I'm hoping you can help me get through this difficult time.}\\
\hline
Supporter &\makecell[l]{Hi, can you please explain the difficult time you are having?} \\
\hline
Seeker & \makecell[l]{My boyfriend broke up with me and I've been upset. We live in different parts of the country and...} \\
\hline
&\makecell[l]{......}\\
\hline
Seeker & \makecell[l]{Have you ever gone through a break up?}\\
\hline
\multicolumn{2}{|c|}{\textbf{Response}}\\
\hline
Transformer & \makecell[l]{I am so sorry that you are not able to get your friends and family. (\textit{Inconsistency})}\\
\hline
MT Transformer & \makecell[l]{I have had a similar situation and I understand. It was the only way I was in similar situation, I have \\ found that way to get out of my life. I have ... and i was in a similar situation. (\textit{Repetition})}\\
\hline
MIME & \makecell[l]{I know it is hard, but it is always good to do things out for yourself. (\textit{Poor Empathy})}\\
\hline
MoEL & \makecell[l]{I am so sorry that you are feeling better. I had a relationship with your boyfriend! (\textit{Aggressiveness})}\\
\hline
BlenderBot-Joint & \makecell[l]{I have not but i have been through a lot of breakups... (\textit{Contradiction})}\\
\hline
MISC  &\makecell[l]{I have been through a breakup and I can tell you that it was very painful. I have had to move on \\ from a relationship because of it. (\textit{Strong Empathy})}\\\hline
\textbf{\emph{Ground-Truth}} & \makecell[l]{It sounds like you are really stressed over the breakup and you want to know how to move forward.}\\
\hline
\end{tabular}
\caption{Responses generated from MISC and other compared models. Some words are omitted due to space limit.}
\label{tab:generate example}       % Give a unique label
\end{table*}

\subsection{Fine-grained Emotion Understanding}
As discussed before, one limitation of previous approaches is that they solely rely on a conversation-level emotion label, which is too coarse to guide the chatbot respond strategically and help the emotional conversation progress healthily. To remedy this issue, we exploit the commonsense knowledge generator COMET to supplement fine-grained information of seeker's mental state. 

In order to fairly examine the effects of different emotional information, we discard the COMET blocks and implement a variant of our method {MISE}, a.k.a. \textbf{MI}xed-\textbf{S}rategy-aware model integrating \textbf{E}motion, where an extra emotion classification objective is added to the main architecture, as in~\citet{rashkin2018know}. Table~\ref{tab:compare-cmb/e} summarizes the comparison results between our full model {MISC} and its variant {MISE}. Obviously, all the metrics drop when replacing the fine-grained mental information with coarse-grained emotion label.

\begin{figure}[!h]
    \centering
    \includegraphics[width=0.45\textwidth]{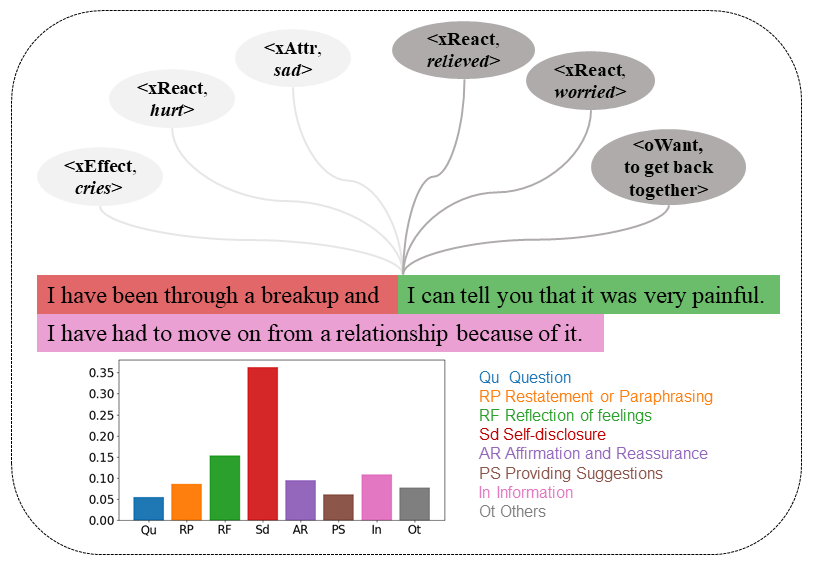}
    \caption{The visualization of how the MISC organizes the response under the effect of multiple factors.} %(The white ovals and grey ovals represent the COMET blocks retrieved using the seeker’s situation information $s$ and last post $x$ respectively.)}
    \label{fig:visualization of the relation}
\end{figure}

To depict the advantage of fine-grained mental state information, we visualize the attended COMET blocks of the example in Table~\ref{tab:generate example}. As shown in Figure~\ref{fig:visualization of the relation}, our chatbot {MISC} pays much attention of those inferred knowledge that are beneficial for fine-grained emotion understanding and strategy-aware empathetic responding.

More specifically, the attended COMET blocks (\verb|xReact|, \emph{hurt}) and (\verb|xAttr|, \emph{sad}) permit our chatbot {MISC} to utter the words ``\emph{it was painful}'' which reflects its understanding of the seeker's feeling. Besides, note that the COMET blocks with white background are retrieved using the situation information $s$, and the grey ones are collected using the seeker's last post $x$. Despite of some overlapping, the white and grey attended blocks do contain distinct and crucial mental state knowledge. This partially validates that $s$ and $x$ is complementary to each other, and they two are useful information for emotional support conversation.

\begin{figure*}[!h]
\centering
\subfigure[predicted by the MISC.]{
\begin{minipage}[t]{0.33\linewidth}
\centering
\includegraphics[width=2in]{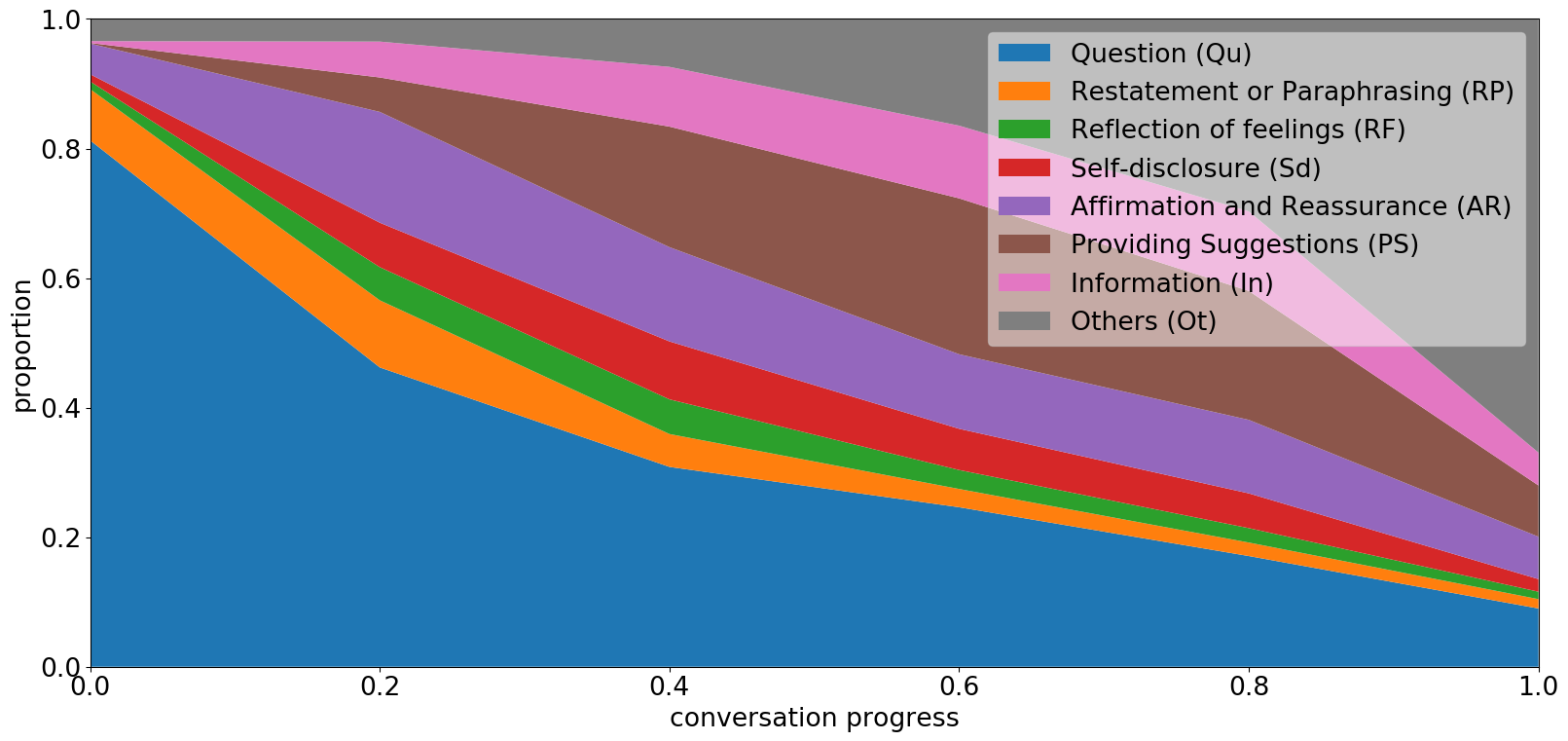}
%\caption{fig1}
\end{minipage}%
}%
\subfigure[from the test set.]{
\begin{minipage}[t]{0.33\linewidth}
\centering
\includegraphics[width=2in]{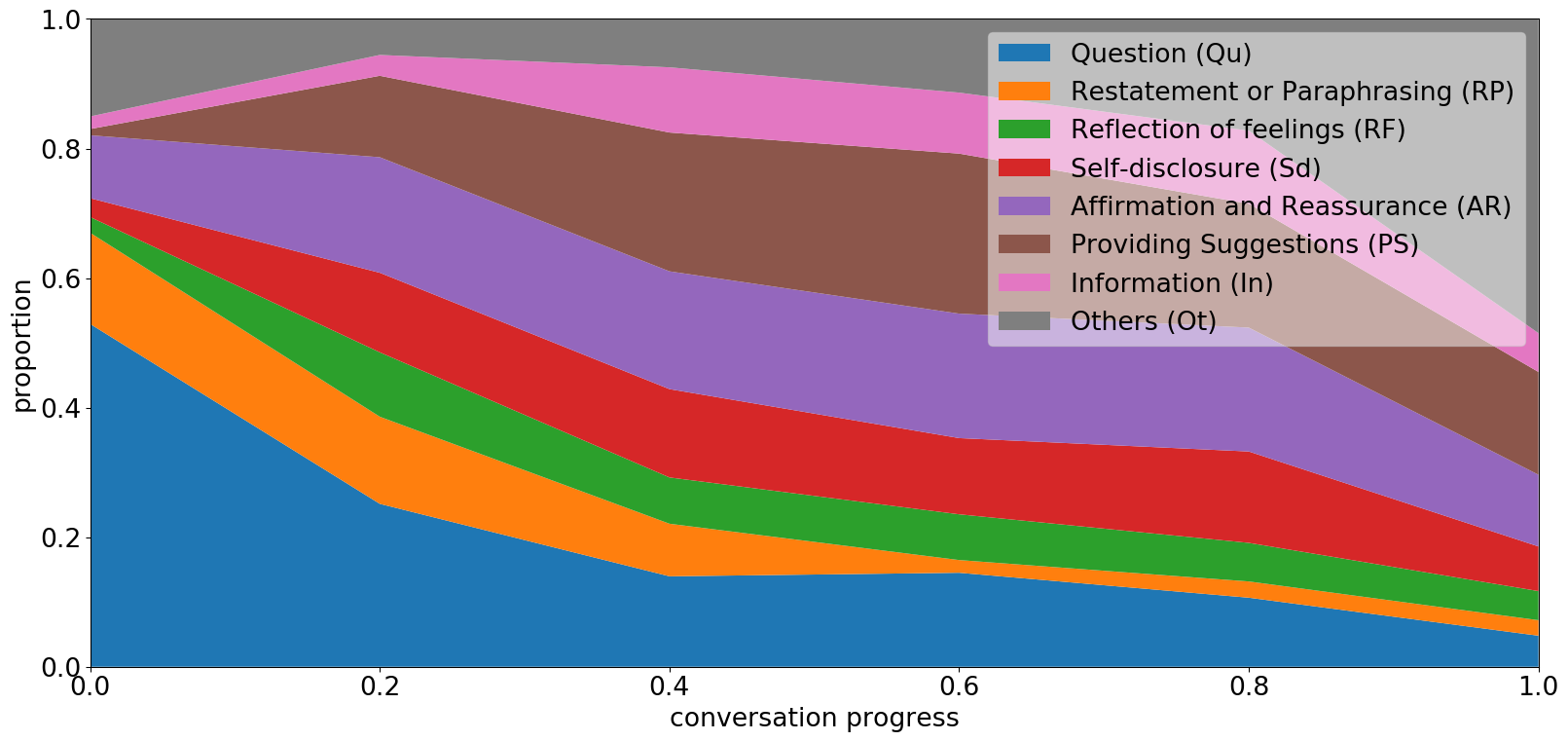}
%\caption{fig2}
\end{minipage}
}%
\subfigure[predicted by the BlenderBot-Joint.]{
\begin{minipage}[t]{0.33\linewidth}
\centering
\includegraphics[width=2in]{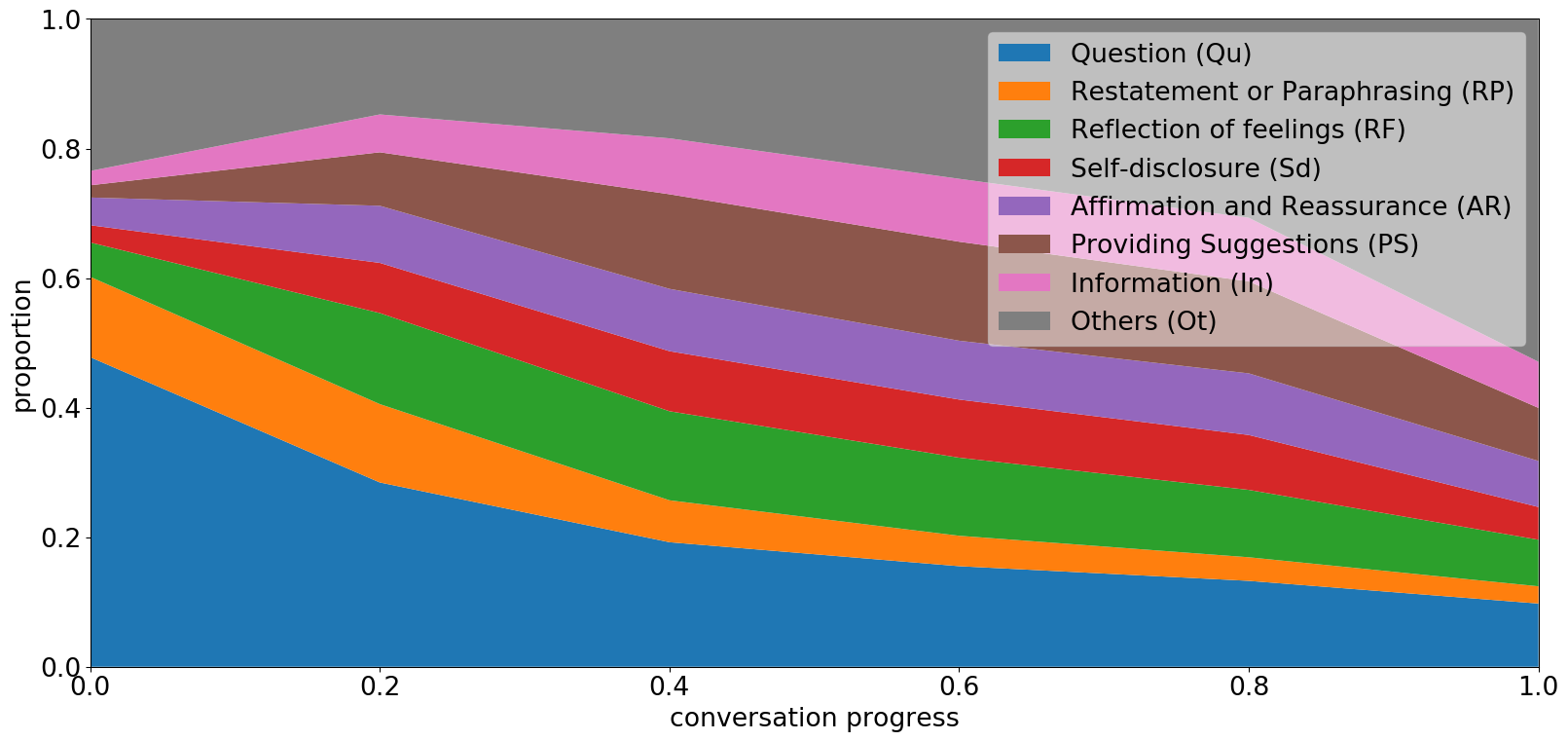}
%\caption{fig2}
\end{minipage}%
}%
\caption{The strategy distribution in the different stage of conversation. }
\label{distribution}
\end{figure*}

\begin{table}[!t]
\centering
\small
%\vspace{-1mm}
\begin{tabular}{lcccc}
\toprule
\textbf{Component}& \textbf{D-1}\,$\uparrow$ &\textbf{B-2}\,$\uparrow$ & \textbf{R-L}\,$\uparrow$&\textbf{M}(\%)\,$\uparrow$\\ 
\midrule
MISC & \textbf{4.41}&\textbf{7.31}&\textbf{17.91}&\textbf{11.05}\\
MISE & 3.94&7.09&16.93&10.53\\
\bottomrule
\end{tabular}
\caption{Results of MISC with Different Emotions.}
\label{tab:compare-cmb/e}
\end{table}

\subsection{Mixed-Strategy-Aware Empathetic Responding}
\label{sec:analysis}
Meanwhile, the mixture of response strategy also plays a vital role for emotional support conversation. 
% In this work, we design a novel mixed representation for strategy and integrate it into response generation. 
By analyzing the aforementioned case in depth, we find some hints on why our way to model conversation strategy is more preferred in the setting of emotional support conversation. 

\noindent\textbf{Hint 1: Mixed strategy is beneficial for Smooth Emotional Support}. In Figure~\ref{fig:visualization of the relation}, we visualize the predicted strategy representation and the generated support response in Table~\ref{tab:generate example}. After understanding the seeker's situation of break-up and feelings of sadness, our {MISC} reasons that it might be proper to employ the strategies of \emph{Self-disclosure}, \emph{Reflection of feelings} to emotionally reply and effectively console the seeker's. Then, {MISC} organizes the response by firstly reveals that ``it'' has similar experiences and knows the feelings like. Moreover, the chatbot also supplements detailed information of \emph{move on from a relationship} to suggest that the life will go on. These added-up words could be regarded as using the strategy of \emph{Information} or \emph{Others}, which is useful to transit the conversation to the next step smoothly. This case vividly shows how response generation is guided by the mixed strategies, and how skillful of our chatbot {MISC} is.
% \footnote{Please refer to the appendix file for detailed definition of these strategies.}
% Please refer to the original paper~\cite{liu2021towards} for detailed definition of these strategies. 

\noindent\textbf{Hint 2: Mixed strategy is more effective than single strategy}. In addition to the case study, we also attempt to quantitatively assess the benefit of the mixed strategy modeling. To do so, we implement another variant of our chatbot \textbf{Single} where the mixed representation is replaced with an one-hot representation. Typically, we pick up the strategy dimension with the largest probability value as the one-hot output. The comparison results are given in Table \ref{tab:compare-msg/ssg}. Although yielding a slightly better distinct-n scores, the single-strategy variant lags far behind according to the lexical and semantic scores. 

\begin{figure}[!h]
    \centering
    \includegraphics[width=0.45\textwidth]{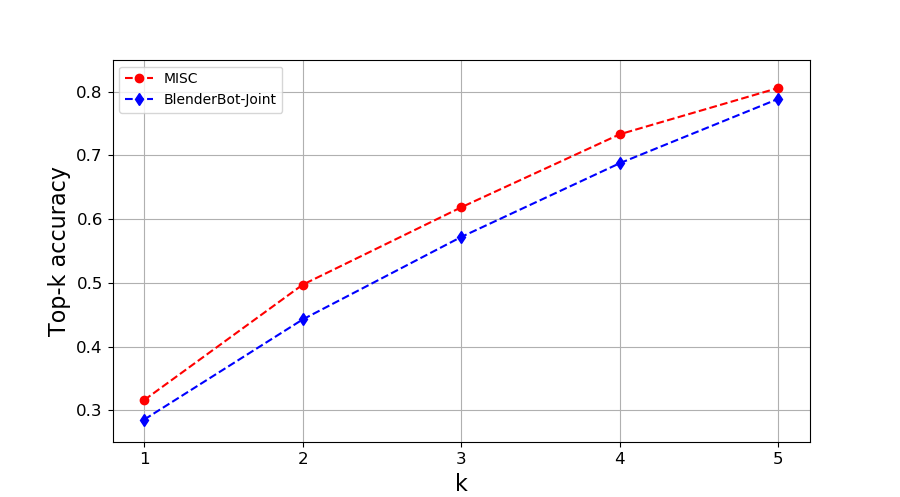}
    \caption{The Top-k Strategy Prediction Accuracy.}
    \label{fig:top-k accuracy}
\end{figure}

Recall that the SOTA model {BlenderBot-Joint}~\cite{liu2021towards} can also be regarded as a single-strategy model where a special strategy token is firstly decoded at the beginning of the response generation. We then compare their way of strategy modeling with our mixed strategy representation. As shown in Figure~\ref{fig:top-k accuracy}, the top-k strategy prediction accuracy of our MISC always surpasses that of BlenderBot-Joint, and the top-5 accuracy of our model reaches over 80\%. This again proves the success of our strategy modeling. %which proves the use of mixed strategy is well-founded.

\begin{table}[htbp]
\centering
\small
%\vspace{-1mm}

\begin{tabular}{lcccc}
\toprule
\textbf{Strategy}& \textbf{D-1}\,$\uparrow$ &\textbf{B-2}\,$\uparrow$ & \textbf{R-L}\,$\uparrow$&\textbf{M}(\%)\,$\uparrow$\\ 
\midrule
Mixture& 4.41&\textbf{7.31}&\textbf{17.91}&\textbf{11.05}\\
Single& \textbf{4.79}&6.30&17.01&10.22\\
\bottomrule
\end{tabular}
\caption{Comparison of different strategy modeling.}
\label{tab:compare-msg/ssg}
\end{table}

\noindent\textbf{Hint 3: Mixed strategy is suitable for ESC Framework}. 
The emotional support conversations in the dataset ESConv are guided by the ESC Framework, which suggests that emotional support generally follows a certain order of strategy flow. %For instance, the strategy of \emph{Question} is often adopted by the supporters to explore more about the seeker's situation, and the strategy of \emph{Providing Suggestions} will be more likely to chosen as the conversation develops. Please refer to the original paper~\cite{liu2021towards} for more details.
Similar to~\cite{liu2021towards}, here we also visualize the strategy distributions learned from different models, and compare them with the ``ground-truth'' strategy distribution in the original dataset. As shown in Figure~\ref{distribution}, we can find: (1) Comparing our model with the SOTA model {BlenderBot-Joint}, we can find that our {MISC} better mimics the skill of strategy adoption in emotional support conversation. (2) At almost all stages of the conversation, our model is less likely to predict the strategy of \emph{Others} (the grey part), as compared to {BlenderBot-Joint}. This indicates that the strategy acquired by our model is more discriminative than those by BlenderBot-Joint. (3) Overall speaking, the strategy distribution from our model share very similar patterns as compared to the ground-truth distribution. This implies that our way to model the strategy learning is suitable for the ESC framework. 
%(2) At the beginning of the conversation, the strategies are prone to \textit{Question} and \textit{Affirmation and Reassurance} to comfort seeker. (3) At the mid stage when the supporter has a basic understanding of the seeker, the major strategy transits to \textit{Providing Suggestions}. (4) At the end of conversation, more \textit{Others} strategy appear, which means that the signal of terminating the conversation lead supporter to choose a few specific response like \textit{bye}, \textit{you are welcome}, etc. On the one hand, these findings confirm that the strategy prediction of the MISC is reasonable. On the other hand, we believe that these findings could instruct both human support and machine support. 
% The computation details are given in Appendix. %The difference is that we use the logical value of strategy to compute proportion rather than the one-hot value.

%% file: 7-conclusion.tex
\section{Conclusions}
In this paper, we propose MISC, a novel framework for emotional support conversation, which introduces COMET to capture user's instant mental state, and devises a mixed strategy-aware decoder to generate supportive response. Through extensive experiments, we prove the superiority and rationality of our model. In the future, we plan to learn the mixed response strategy in a dynamic way.

\section{Ethical Considerations}
At last, we discuss the potential ethic impacts of this work: (1) The ESConv dataset is a publicly-available, well-established benchmark for emotional support conversation; % The research purpose is to analyze how to strategically reduce other's distress through turns of empathetic conversations; 
(2) \textbf{Privacy}: The original providers have filtered the sensitive information such as personally identifiable information~\cite{liu2021towards}; (3) Nevertheless, due to the limitation of filtering coverage, the conversations might still remain some languages that are emotionally triggering. Note that our work focuses on building emotional support conversational agents. For risky situations such as self-harm-related conversations, we do not claim any treatments or diagnosis.